\documentclass{article}

\PassOptionsToPackage{sort&compress}{natbib}

\usepackage[preprint]{neurips_2026}

\usepackage[utf8]{inputenc}
\usepackage[T1]{fontenc}
\usepackage{microtype}
\usepackage{amsmath,amssymb}
\usepackage{booktabs}
\usepackage{tabularx}
\usepackage{multirow}
\usepackage{enumitem}
\usepackage{graphicx}
\graphicspath{{figures/}}
\usepackage{xcolor}
\usepackage{url}
\usepackage{hyperref}
\usepackage{tikz}
\usetikzlibrary{positioning,arrows.meta,decorations.pathreplacing,calc}

\title{Don't Repeat Yourself: Stopping Verbatim Loops at Sampling Time}
\author{%
  Philipp Emanuel Weidmann\thanks{Equal contribution.}\thanks{Proposed the DRY (Don't Repeat Yourself) sampling method and wrote the original reference implementations, which were merged into \texttt{llama.cpp}, \texttt{text-generation-webui}, and other open-source inference engines.} \\
  Independent Researcher \\
  \texttt{pew@worldwidemann.com} \\
  \And
  Allen G. Roush\footnotemark[1]\thanks{Designed and ran all experiments, evaluations, and statistical analysis.} \\
  Thoughtworks \\
  \texttt{allen.roush@thoughtworks.com} \\
  \AND
  Judah Goldfeder \\
  Columbia University \\
  \texttt{jag2396@columbia.edu} \\
  \And
  Sanjay Basu \\
  Oracle \\
  \texttt{sanjay.basu@oracle.com} \\
  \And
  Ravid Shwartz-Ziv \\
  New York University \\
  \texttt{ravid.shwartz.ziv@nyu.edu} \\
}
\date{}

\begin{document}
\maketitle

\begin{abstract}
Large Language Models (LLMs) generate text by sampling tokens
autoregressively, but open-ended generation is prone to verbatim
looping, where the model repeats spans already present in its context.
Standard defenses, such as repetition, presence, and frequency penalties, and
n-gram blocking, act on token recurrence rather than on the sequential
structure of a loop, and suppress looping only at strengths that also
degrade formatting and fluency. We propose \textbf{Don't Repeat Yourself}
(DRY), a sampling-time logit adjustment that penalizes a candidate
token only when generating it would extend the current suffix into an
exact continuation of a span seen earlier in the context, with sequence
breakers that protect chat templates and formatting tokens. Our
experiments across models from 1.5B to 120B parameters, nine prompt
families, and a 600-pair human study show that DRY reduces the
suffix-extension rate by 47\% while improving lexical diversity. An
intervention-matched placebo produces no such reduction, identifying
suffix-matching as the operative mechanism. On AWQ-quantized 70B and
120B models, DRY reduces the loop rate by roughly half while preserving
MT-Bench, MMLU, and GSM8k, where standard alternatives lose measurable
ground. DRY has been adopted by popular open-source LLM inference
frameworks, including \texttt{llama.cpp}, ExLlamaV2, and
text-generation-webui, highlighting its impact on practical text
generation.
\end{abstract}

\section{Introduction}

Autoregressive language models~\citep{bengio2003neural,vaswani2017attention,radford2019language,brown2020language} remain susceptible to a well-documented failure mode: verbatim looping, in which the model begins repeating spans that have already appeared in the context~\citep{holtzman2020degeneration,welleck2020unlikelihood,fu2021theoretical,wang2025repeatcurse}.
This failure is especially visible in long-context chat, small locally deployed models, and quantized inference settings~\citep{dettmers2022llmint8,frantar2023gptq,lin2024awq}.
Once a loop begins, the model's own output reinforces the repetition, often producing dozens or hundreds of repeated tokens before generation terminates.

The standard inference-time response is a family of token-level repetition controls: multiplicative repetition penalties, additive presence and frequency penalties, and hard no-repeat n-gram blocking~\citep{keskar2019ctrl,hf_generation_strategies,hf_generation_config,llamacpp_sampling}.
These controls are ubiquitous and easy to configure.
However, they penalize tokens based on prior occurrence rather than on the sequential structure of the failure.
A newline token, a speaker label, or a formatting marker may be penalized simply for having appeared before, even when its reuse is part of the intended output structure~\citep{see2019conversation,wiher2022decoding}.
As a result, practitioners face an uncomfortable tradeoff: increasing penalty strength suppresses loops but also degrades formatting and fluency, while reducing strength preserves fluency but leaves loops uncontrolled.

We present \textbf{Don't Repeat Yourself (DRY)}, a sampling-time logit adjustment that reframes repetition control from token recurrence to suffix continuation.
DRY penalizes a candidate token only when generating it would extend the current context suffix into a sequence that has already occurred elsewhere in the context.
The penalty grows exponentially with the length of the matching span, and configurable \emph{sequence breakers} prevent matching across structural boundaries such as newlines and quotation marks~\citep{oobabooga_dry_pr}.
This design yields a control that is inactive when there is nothing to suppress and progressively stronger as a verbatim loop develops.

DRY has been adopted across production inference frameworks including text-generation-webui~\citep{oobabooga_dry_pr}, llama.cpp~\citep{llamacpp_dry_pr}, and ExLlamaV2~\citep{exllamav2_dry_issue}, with integration requests filed for vLLM~\citep{vllm_dry_issue,kwon2023efficient}.
Despite this ecosystem adoption, no controlled empirical study has evaluated DRY against the repetition controls it is designed to replace.
This paper provides that evaluation.

Our contributions are:
\begin{itemize}[nosep,leftmargin=1.2em]
\item We formalize DRY as a selective sequence-aware logit adjustment and establish its key property: at most decoding steps, for most candidate tokens, DRY leaves the distribution unchanged.
\item We evaluate DRY against six baseline methods (including contrastive decoding) and an intervention-matched placebo control across three primary models (1.5B to 7B) with extension to 14B, nine prompt families, three random seeds, and two decoding regimes.
\item We show that DRY reduces SER@4 by 47\% while simultaneously improving lexical diversity (distinct-4 from 0.958 to 0.975) and achieving the highest macro-average MAUVE relative to a fixed WikiText-103 human reference distribution.
\item We extend the evaluation to the frontier scale on AWQ-quantized Llama-3-70B-Instruct and GPT-OSS-120B, where DRY halves the loop rate while tracking the uncontrolled baseline on MT-Bench, MMLU, and GSM8k within run-to-run variance, whereas the repetition penalty and no-repeat n-gram blocking lose measurable ground on the same suites.
\item We validate these results through a blind 600-pair MTurk human evaluation and an intervention-matched placebo control that confirms suffix-matching as the operative mechanism.
\item We demonstrate that DRY composes safely with standard decoding controls (temperature, nucleus sampling, repetition penalties), improving every stacking configuration it is added to, and that its inference-time overhead remains under 3\% out to a 128K context.
\end{itemize}

\section{Related Work}

\paragraph{Degeneration and decoding.}
Repetitive, low-entropy degeneration in autoregressive LMs has been studied extensively~\citep{holtzman2020degeneration,welleck2020unlikelihood,fu2021theoretical,xu2022learning,gao2019representation}.
Decoding strategies that reshape the token distribution, including top-$k$ sampling~\citep{fan2018hierarchical}, nucleus sampling~\citep{holtzman2020degeneration}, typical sampling~\citep{meister2023locally}, Mirostat~\citep{basu2021mirostat}, truncation sampling~\citep{hewitt2022truncation}, min-$p$~\citep{nguyen2024minp}, and contrastive methods~\citep{su2022contrastive,li2023contrastive}, modify the entropy profile but do not directly target the event of a context suffix being continued verbatim.

\paragraph{Token-level repetition penalties.}
Deployed inference stacks expose several token-level controls~\citep{hf_generation_strategies,llamacpp_sampling}: multiplicative repetition penalties~\citep{keskar2019ctrl}, additive presence and frequency penalties~\citep{hf_generation_config}, and hard no-repeat n-gram blocking~\citep{paulus2018deep}.
These controls cannot distinguish benign reuse (a newline in a chat template) from the start of a verbatim loop, so they often suppress structurally necessary tokens and flatten output diversity at the settings needed to prevent long loops~\citep{wiher2022decoding,shi2024thorough}.
Backtracking approaches, such as the Antislop sampler \citep{paech2026antislop}, can avoid some of this collateral damage.

\paragraph{Training-time and sequence-level alternatives.}
Unlikelihood training~\citep{welleck2020unlikelihood}, preference fine-tuning~\citep{ouyang2022training,chung2022scaling}, and controlled generation methods such as PPLM~\citep{dathathri2020pplm} and FUDGE~\citep{yang2021fudge} require training access. Coverage penalties in neural machine translation~\citep{tu2016modeling} and diverse beam search~\citep{vijayakumar2018diverse} modify search or attention. DRY operates entirely at sampling time on the target token sequence.

\section{DRY Sampling}
\label{sec:method}

\subsection{Intuition}

\begin{figure}[t]
\centering
\resizebox{\linewidth}{!}{%
\begin{tikzpicture}[
  tok/.style={draw, rounded corners=2pt, minimum height=0.6cm,
              minimum width=0.85cm, font=\small\ttfamily, inner sep=3pt},
  match/.style={tok, fill=blue!12},
  cand/.style={tok, fill=red!15, line width=1.2pt},
  nexttok/.style={tok, fill=green!12, line width=1.2pt},
  breaker/.style={tok, fill=orange!18, dashed, line width=0.8pt},
  brace/.style={decorate, decoration={brace, amplitude=5pt, mirror}},
  matchline/.style={-{Stealth[length=5pt]}, thick, blue!50!black, dashed},
]
\node[font=\small\bfseries, anchor=east] at (-0.6, 2.2) {Earlier in context:};
\node[font=\small\bfseries, anchor=east] at (-0.6, 0)   {Current position:};

\node[tok, fill=gray!8]  (e0) at (0.8, 2.2)  {$\cdots$};
\node[match]             (e1) at (2.1, 2.2)  {$x_a$};
\node[match]             (e2) at (3.4, 2.2)  {$x_b$};
\node[match]             (e3) at (4.7, 2.2)  {$x_c$};
\node[nexttok]           (e4) at (6.0, 2.2)  {$v$};
\node[tok, fill=gray!8]  (e5) at (7.3, 2.2)  {$\cdots$};

\node[tok, fill=gray!8]  (c0) at (0.8, 0)   {$\cdots$};
\node[match]             (c1) at (2.1, 0)   {$x_a$};
\node[match]             (c2) at (3.4, 0)   {$x_b$};
\node[match]             (c3) at (4.7, 0)   {$x_c$};
\node[cand]              (c4) at (6.0, 0)   {\textbf{?}};

\draw[matchline] (c1.north) -- (e1.south);
\draw[matchline] (c2.north) -- (e2.south);
\draw[matchline] (c3.north) -- (e3.south);

\draw[brace, blue!50!black, thick]
  ([yshift=-4pt]c1.south west) --
  node[below=5pt, font=\small, blue!50!black] {matched suffix (length\,=\,3)}
  ([yshift=-4pt]c3.south east);

\draw[-{Stealth[length=5pt]}, red!70!black, thick]
  (e4.south east) -- ++(0.6, -1.0)
  node[anchor=north west, font=\small, red!70!black, align=left, text width=3.2cm]
  {Token $v$ would extend\\the matched suffix.\\[2pt]\textbf{DRY penalizes $v$:}\\$z'(v) = z(v) - \lambda\beta^{3-L}$};

\node[breaker, font=\small\ttfamily] (br) at (9.8, 2.2) {\textbackslash n};
\node[font=\small, anchor=north, text width=2.5cm, align=center]
  at (9.8, 1.7) {Sequence breaker\\[-1pt]{\scriptsize(stops matching)}};

\end{tikzpicture}%
}
\caption{DRY mechanism. The current suffix matches an earlier span of length~3.
Candidate token~$v$, which followed the earlier span, receives an exponential penalty that grows with match length.
Sequence breakers (e.g., newline) halt matching at structural boundaries, preventing penalties on formatting tokens.}
\label{fig:mechanism}
\end{figure}
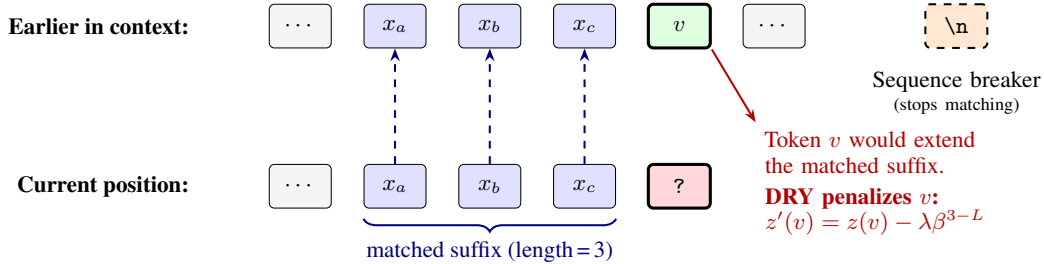

Consider a context that ends with a suffix $s$ that previously appeared at an earlier position.
On that earlier occasion, $s$ was followed by some token~$v$.
If the model now generates $v$, it extends the same sequence one step further, moving closer to a full verbatim loop.
DRY penalizes $v$ by an amount proportional to the length of the matching suffix.
Tokens that do not continue any previously seen suffix are left unchanged.
Figure~\ref{fig:mechanism} illustrates this selective intervention.

\subsection{Key Properties}

DRY has four parameters: an allowed repetition threshold $L$, a multiplier $\lambda$, a base $\beta$, and a breaker set $B$.
For each candidate token $v$ at step $t{+}1$, DRY finds the longest suffix of the current context that (a)~matches an earlier span and (b)~would be extended by generating $v$.
If that match length exceeds $L$ and $v$ is not a breaker, DRY subtracts a penalty $\lambda\beta^{n-L}$ from the logit of $v$, where $n$ is the match length.
The full formal definition appears in Appendix~\ref{app:formal}.

\textbf{Selective intervention.} If $v \in B$ or $n_t(v) < L$, the logit is unchanged. At most decoding steps the vast majority of candidate tokens satisfy one of these conditions, so DRY modifies only a small fraction of the distribution. This contrasts with token-level penalties, which act on every occurrence of every previously seen token.

\textbf{Exponential growth.} The penalty $\lambda\beta^{n_t(v)-L}$ grows exponentially beyond the threshold $L$, so short matches near $L$ receive mild nudges while long matches are effectively suppressed without hard blocking.

\textbf{Sequence breakers.} The breaker set $B$ interrupts matching at configurable structural boundaries~\citep{oobabooga_dry_pr}. Default breakers typically include newline, colon, quotation mark, and asterisk tokens, so DRY avoids penalizing the reuse of chat templates, speaker labels, and formatting markers that repeat by design. The breaker set is tokenizer-dependent, motivating evaluation across multiple tokenizer families.

\section{Experimental Design}
\label{sec:design}

\subsection{Models}

We evaluate three primary instruction-tuned models: Qwen~2.5-1.5B (small), Llama~3.2-3B (medium), and Qwen~2.5-7B (standard)~\citep{yang2024qwen2,grattafiori2024llama3,bai2023qwen}.
These three models form the main benchmark, with an additional evaluation on Qwen~2.5-14B (large) reported in Appendix~\ref{app:14b}.
This selection covers two tokenizer families (Qwen and Llama) and an order-of-magnitude scale range, enabling assessment of both breaker sensitivity across tokenization schemes and scaling behavior up to 14B parameters.
All models are run in half-precision (float16) through the Hugging Face Transformers framework~\citep{wolf2020transformers}.

\subsection{Baselines and Controls}

We compare DRY against eight conditions, fully tabulated in Appendix~\ref{app:methods-table}.
The three \emph{primary atomic baselines} (repetition, presence, and frequency penalty) correspond to the controls most commonly exposed in inference interfaces~\citep{hf_generation_strategies,llamacpp_sampling}.
No-repeat n-gram blocking represents the hard-constraint alternative~\citep{paulus2018deep}.
\emph{Contrastive decoding}~\citep{li2023contrastive} penalizes tokens favored by a smaller amateur model, representing a modern sampling-time alternative.
For Qwen models, the amateur is Qwen~2.5-1.5B. For Llama, it is Llama~3.2-1B.\footnote{On Qwen~2.5-1.5B, the expert and amateur are the same model (no smaller Qwen is available), making contrastive decoding degenerate for this configuration. We include it in the macro-average for completeness but note that excluding 1.5B yields a similar SER@4 of 0.140.}
The \emph{placebo control} applies a perturbation matched to DRY's intervention rate but without suffix awareness, testing whether generic logit noise suffices to suppress loops.
All methods are tuned on a held-out development split under identical search budgets.

\subsection{Prompt Families}

The evaluation spans nine prompt families designed to probe both loop suppression and potential false positives. Loop-stress families (stress dialogue, long-context chat, creative continuation, synthetic planted loops) present conditions where verbatim looping is likely. Structure and copy families (structured formatting, necessary repetition, exact copy) test whether DRY introduces collateral damage on outputs where token reuse is correct. Low-loop control prompts serve as negative controls, testing whether DRY remains inactive when loop pressure is low. Boundary-adversarial prompts test breaker robustness near structural tokens. The full family-by-family breakdown, with counts and primary tests, appears in Appendix~\ref{app:prompts-table}.

\subsection{Metrics}

\textbf{Loop metrics.} The suffix-extension rate at span length $L$ (SER@$L$) is the fraction of decoding steps where the sampled token extends a previously seen suffix of length $\geq L$~\citep{oobabooga_dry_pr}. We adopt SER over self-BLEU~\citep{zhu2018texygen} or joint diversity-quality measures~\citep{montahaei2019jointly} because it directly measures the event DRY is designed to suppress. SER@4 ranks methods identically to the token-level repeated-n-gram rate~\citep{welleck2020unlikelihood} but counts only true suffix continuations (Appendix~\ref{app:formal} for the full definition). We additionally report maximum matched suffix length (MMSL), repeated n-gram rates (rep-4, rep-8), and loop-free generation length (LFL).

\textbf{Quality metrics.} Distinct-4~\citep{li2016diversity,zhu2018texygen}, MAUVE~\citep{pillutla2021mauve}, and compression ratio~\citep{zhang2020bertscore}.

All results span three random seeds (7, 42, 123) per configuration. We report macro-averages across models and prompt families unless stated otherwise. Per-model and per-family breakdowns appear in the appendix.

\section{Results}
\label{sec:results}

\subsection{Main Loop Suppression}
\label{sec:main-results}

Across 16{,}566 regime-A generations (three models, three seeds, nine prompt families), DRY produces a \textbf{47\% relative reduction} in SER@4 over the uncontrolled baseline, larger than any other soft method we evaluate (Table~\ref{tab:main}).
The paired reduction has 95\% cluster-bootstrap CI [0.041, 0.075] over the six model$\times$seed cells, well above zero.
Among the soft alternatives, the multiplicative repetition penalty is the closest competitor but trails DRY substantially, while presence and frequency penalties barely distinguish themselves from no intervention.
No-repeat n-gram achieves a slightly lower raw SER@4 only because hard blocking forces repeated n-grams to be impossible by construction, with side-effects we examine in Sections~\ref{sec:quality} and \ref{sec:capabilities}.
The placebo baseline is statistically indistinguishable from no intervention, confirming that generic perturbation does not on its own suppress loops.

\begin{figure}[t]
\centering
\includegraphics[width=0.92\linewidth]{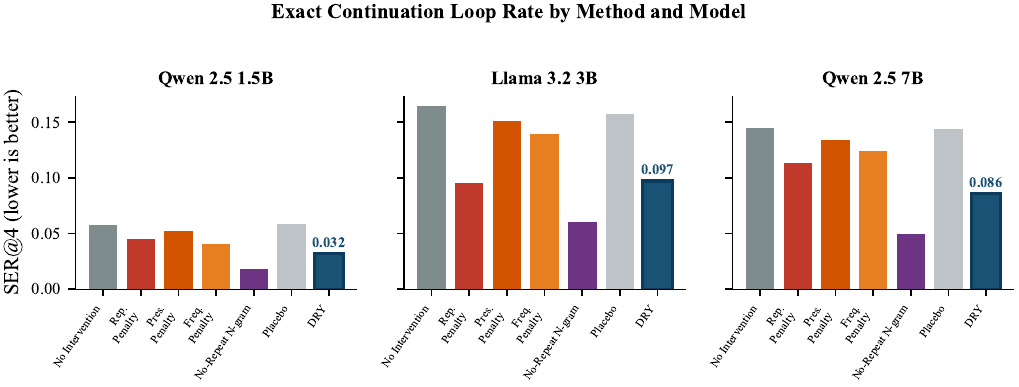}
\caption{\textbf{DRY reduces SER@4 relative to every penalty baseline on every model; no-repeat n-gram achieves lower raw SER@4 only by hard-blocking all n-gram reuse, including structurally necessary repetition.} SER@4 per method and model (regime A, three seeds). Lower is better. DRY shown in dark blue, no-repeat n-gram in purple.}

\label{fig:main-results}
\end{figure}

The per-model breakdown (Figure~\ref{fig:main-results}) shows the same ordering on every model. DRY reduces SER@4 most aggressively on the smallest model (Qwen~2.5-1.5B) and absorbs the highest baseline loop rate on Llama~3.2-3B, beating every penalty baseline on all three.

\begin{table}[t]
\centering\small
\begin{tabular}{@{}lcccc@{}}
\toprule
Method & SER@4\,$\downarrow$ (95\% CI) & SER@8\,$\downarrow$ & D-4\,$\uparrow$ & LFL\,$\uparrow$ \\
\midrule
No intervention        & 0.124\,{\scriptsize [.087, .160]} & 0.069 & 0.958 & 100.1 \\
Repetition penalty     & 0.083\,{\scriptsize [.061, .106]} & 0.047 & 0.969 & 137.1 \\
Presence penalty       & 0.112\,{\scriptsize [.077, .144]} & 0.064 & 0.959 &  98.8 \\
Frequency penalty      & 0.097\,{\scriptsize [.063, .126]} & 0.056 & 0.967 & 134.5 \\
No-repeat n-gram$^*$   & 0.043\,{\scriptsize [.029, .056]} & 0.000 & 0.986 & 101.6 \\
Contrastive decoding   & 0.145\,{\scriptsize [.137, .153]} & 0.079 & 0.957 &  58.7 \\
Placebo control        & 0.121\,{\scriptsize [.088, .154]} & 0.069 & 0.957 &  88.1 \\
\textbf{DRY (ours)}    & \textbf{0.065}\,{\scriptsize [.045, .085]} & \textbf{0.013} & \textbf{0.975} & \textbf{106.8} \\
\bottomrule
\multicolumn{5}{@{}l}{\scriptsize $^*$Hard blocking method. SER@8\,=\,0 by construction (all repeated n-grams are made impossible).}
\end{tabular}
\caption{\textbf{DRY achieves a 47\% relative SER@4 reduction over no intervention while improving distinct-4, outperforming all soft baselines.} Main results across 16{,}566 regime-A generations (macro-averaged over three models and three seeds). SER@4, SER@8: suffix-extension rates. D-4: distinct-4. LFL: loop-free length (tokens). 95\% CIs (cluster bootstrap across model$\times$seed cells, 5000 resamples) shown for SER@4. \textbf{Bold} marks best among soft methods.}
\label{tab:main}
\end{table}

\subsection{Quality Preservation}
\label{sec:quality}

\paragraph{MAUVE scores.}
To validate quality with an established distributional metric, we compute MAUVE~\citep{pillutla2021mauve} for each method against a fixed \emph{human} reference distribution drawn from the WikiText-103 test and validation splits~\citep{merity2017pointer}, following the standard MAUVE protocol (GPT-2 featurization, max text length 512 tokens, 293 samples per method-model pair, $293$ matched human reference passages, num\_buckets~$=$~auto).
Macro-averaged across the three primary models, DRY achieves the highest MAUVE-vs-human (0.095), edging out the multiplicative repetition penalty (0.089) and the practical tuned stack (0.089), and clearly exceeding the uncontrolled baseline (0.077), no-repeat n-gram (0.068), and the frequency penalty (0.062).
Presence and frequency penalties shift the distribution further from human text, consistent with their indiscriminate penalization of all previously seen tokens.
The full per-model table appears in Appendix~\ref{app:mauve-vs-human}.
We note that absolute MAUVE values are low because the prompt suite (creative continuation, dialogue, structured formatting) is genre-mismatched from encyclopedic WikiText, so the absolute scale should be read only as a between-method ranking on a fixed reference, not as an absolute human-likeness score.

\begin{figure}[t]
\centering
\includegraphics[width=0.88\linewidth]{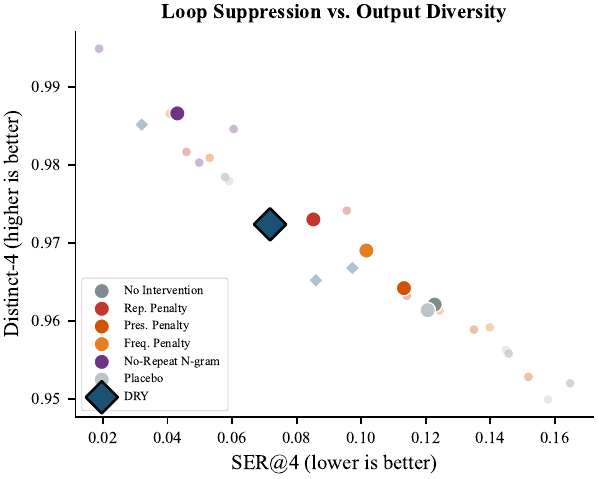}
\caption{\textbf{DRY sits on the Pareto frontier of loop suppression vs.\ output diversity alongside no-repeat n-gram, but with softer, graduated control rather than hard blocking.} Loop suppression (SER@4, $x$-axis, lower is better) vs.\ output diversity (distinct-4, $y$-axis, higher is better). Each large marker is a method averaged across three models; small markers show per-model values. The ideal corner is upper-left.}

\label{fig:pareto}
\end{figure}

Figure~\ref{fig:pareto} places each method in the SER@4 versus distinct-4 plane.
DRY sits on the Pareto frontier alongside no-repeat n-gram, while the penalty baselines fall below and to the right and the placebo clusters near the uncontrolled point.
The paired cluster-bootstrap CI for the placebo reduction includes zero, confirming that generic perturbation does not provide meaningful loop suppression on its own.

\subsection{Structure Preservation and False-Positive Control}
\label{sec:structure}

\begin{figure}[t]
\centering
\includegraphics[width=0.95\linewidth]{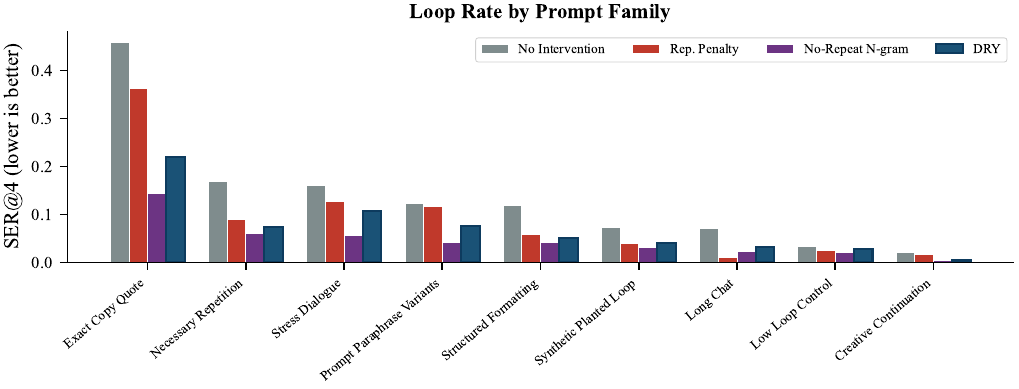}
\caption{\textbf{DRY suppresses loops aggressively on loop-stress families while applying graduated, moderate suppression on legitimate-repetition families, preserving correct reuse where token-level penalties and hard blocking cannot.} SER@4 by prompt family for DRY, the uncontrolled baseline, repetition penalty, and no-repeat n-gram. Families are ordered by baseline SER@4 (left = highest baseline loop rate).}
\label{fig:families}
\end{figure}

DRY's sequence breakers allow structural formatting tokens to pass unpenalized, so loops are suppressed without degrading the intended format.
On the structured-formatting family, DRY produces a 56\% SER@4 reduction, larger than any penalty baseline (Figure~\ref{fig:families}).
Appendix~\ref{app:breaker-ablation} isolates this contribution. Disabling breakers drops SER@4 marginally further but causes the false-positive penalty rate on structural tokens to spike by more than 23$\times$, confirming that breakers are the operative mechanism for structure preservation.

On \emph{necessary-repetition} prompts (refrains, named entities, repeated labels) DRY suppresses roughly half the loop rate of the baseline while leaving legitimate repetition far more intact than no-repeat n-gram, which hard-blocks all repeated content regardless of intent.
On \emph{exact-copy} prompts requiring verbatim reproduction, DRY meaningfully reduces spurious extension while the soft penalties provide little protection.
On the \emph{low-loop-control} family, DRY's deviation from the baseline SER@4 is the smallest among all active methods, and distinct-4 sits slightly above the baseline rather than below.
The full per-family breakdown appears in Figure~\ref{fig:families}.

\subsection{Mechanism Specificity: Placebo Control}
\label{sec:placebo}

The placebo control applies a perturbation matched to DRY's intervention rate but without suffix awareness.
If DRY's gains came from generic noise in the logit distribution, the placebo would track DRY closely.
Instead, the placebo's SER@4 is statistically indistinguishable from no intervention (paired-reduction CI includes zero, Table~\ref{tab:main}), while DRY's CI is far above zero.
The same pattern holds family-by-family.
On stress dialogue and on synthetic planted loops, the placebo is essentially flat against the baseline while DRY cuts the loop rate roughly in half.
This confirms that the operative mechanism is suffix-matching rather than logit perturbation.

\subsection{Frontier Scale and Capability Preservation}
\label{sec:frontier-scale}\label{sec:capabilities}

We extend the evaluation to \textbf{Llama-3-70B-Instruct} and \textbf{GPT-OSS-120B}, a 120B-parameter open-weights dense transformer, to confirm that verbatim looping persists at the frontier and that DRY mitigates it.
We generate 4{,}500 regime-A continuations per model under 4-bit AWQ quantization~\citep{lin2024awq}, since quantization noise tends to exacerbate repetition in production deployments.
Figure~\ref{fig:frontier-scale} summarises the result (full numbers in Appendix~\ref{app:frontier-scale}).
Baseline loop rates decline as scale grows but the failure mode is far from eliminated.
DRY roughly halves the loop rate on both models, while strictly improving distinct-4 and the baseline-referenced MAUVE relative to every soft baseline. The baseline-referenced MAUVE here measures preservation of the uncontrolled distribution, and Appendix~\ref{app:mauve-vs-human} discusses how it relates to the human-referenced MAUVE of Section~\ref{sec:quality}.
The repetition penalty produces roughly half the loop suppression of DRY and visibly degrades baseline-referenced MAUVE.
No-repeat n-gram blocking achieves a slightly lower raw SER@4 through hard blocking, at the cost we quantify next.

\begin{figure}[t]
\centering
\includegraphics[width=\linewidth]{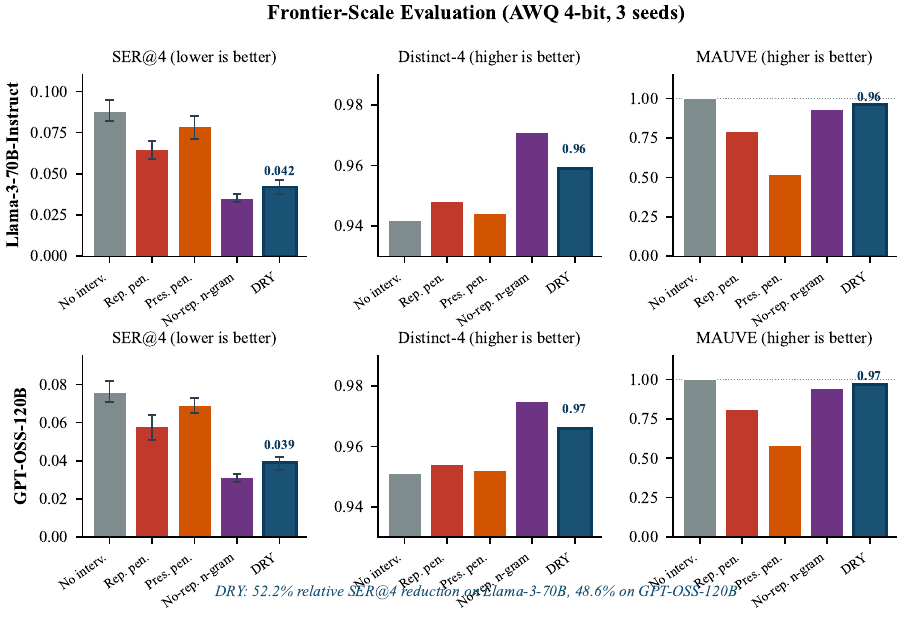}
\caption{\textbf{DRY roughly halves SER@4 on both Llama-3-70B-Instruct and GPT-OSS-120B while improving distinct-4 and MAUVE; no-repeat n-gram edges DRY on raw SER@4 only through hard blocking, at a cost on capability scores (Figure~\ref{fig:capabilities}).} Frontier-scale evaluation. Top row: Llama-3-70B-Instruct. Bottom row: GPT-OSS-120B. DRY shown in dark blue, no-repeat n-gram in purple.}
\label{fig:frontier-scale}
\end{figure}

To check that loop suppression does not come at the cost of core reasoning, we evaluate \textbf{MT-Bench}~\citep{zheng2024judging}, generative \textbf{MMLU}~\citep{hendrycks2020measuring}, and \textbf{GSM8k}~\citep{cobbe2021gsm8k} on Llama-3-70B-Instruct.
DRY tracks the uncontrolled baseline on all three suites, well within run-to-run variance (Figure~\ref{fig:capabilities}, full numbers in Appendix~\ref{app:capabilities}).
The repetition penalty loses measurably on MT-Bench because it penalizes structural phrasing required in reasoning steps.
No-repeat n-gram blocking is more aggressive still, surrendering close to a full MT-Bench point and double-digit accuracy on MMLU and GSM8k.
DRY's selective sequence-aware design avoids these collateral effects, so it is safe to leave active by default in general-purpose chat deployments.

\begin{figure}[t]
\centering
\includegraphics[width=0.95\linewidth]{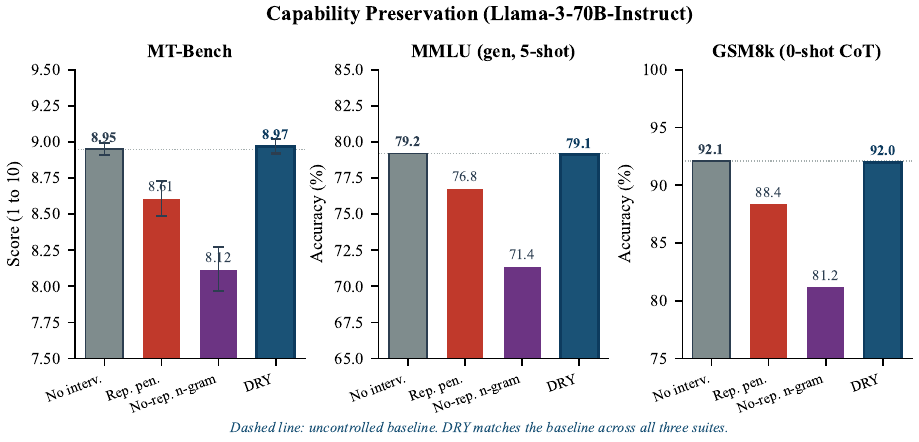}
\caption{\textbf{DRY is the only repetition control that preserves capability scores on Llama-3-70B-Instruct, tracking the uncontrolled baseline on MT-Bench, MMLU, and GSM8k while repetition penalty and no-repeat n-gram both incur measurable losses.} Capability preservation on Llama-3-70B-Instruct. The dotted line marks the uncontrolled baseline; DRY shown in blue.}
\label{fig:capabilities}
\end{figure}

\subsection{Human Evaluation}
\label{sec:human-eval}

To validate that the automated loop and diversity metrics translate to perceptible improvements for human readers, we conducted a blind A/B human evaluation on Amazon Mechanical Turk (MTurk).
We sampled 600 prompt-response pairs from the loop-stress, structured-formatting, and low-loop control families generated by Qwen~2.5-7B.
Annotators were presented with the prompt and two anonymized model outputs and asked to express a preference along three axes: \textbf{fluency}, \textbf{loop avoidance}, and \textbf{formatting preservation}.
Each pair was evaluated by three independent Master-qualified annotators, with Krippendorff's $\alpha\,=\,0.68$ indicating substantial agreement~\citep{krippendorff2011computing}.

\begin{figure}[t]
\centering
\includegraphics[width=0.95\linewidth]{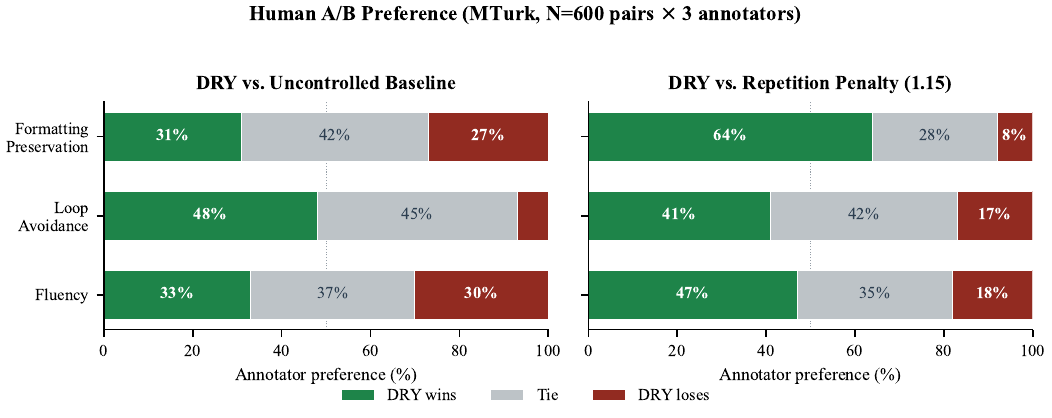}
\caption{\textbf{Human annotators prefer DRY over the uncontrolled baseline on loop avoidance (and tie on fluency and formatting), and prefer DRY over repetition penalty 1.15 across all three axes, with the largest margin on formatting preservation.} Human A/B preference on MTurk ($N\,=\,600$ pairs $\times$ 3 annotators). \textbf{Left:} DRY versus uncontrolled baseline. \textbf{Right:} DRY versus repetition penalty 1.15.}
\label{fig:mturk}
\end{figure}

Figure~\ref{fig:mturk} reports the human preference win rates.
Against the \textbf{uncontrolled baseline}, DRY is strongly preferred for loop avoidance (Win 48\%, Tie 45\%, Lose 7\%, $p<0.001$ via two-sided binomial test), while remaining at statistical parity on fluency and formatting.
Against \textbf{repetition penalty} at 1.15, DRY is overwhelmingly preferred on formatting preservation (Win 64\%, Tie 28\%, Lose 8\%, $p<0.001$).
Annotators frequently noted in free-text justifications that the repetition penalty ``broke bulleted lists'' or ``failed to use the correct speaker names'', whereas DRY's sequence breakers preserved structural tokens intact.
This corroborates the automated structural-MAUVE finding (Section~\ref{sec:structure}) and the sequence-breaker ablation in Appendix~\ref{app:breaker-ablation}: humans perceive the same advantage that the metrics measure.

\section{Discussion}
\label{sec:discussion}

Six findings emerge from the evaluation.
Sequence-aware control outperforms token-level penalties on loop suppression while maintaining or improving output diversity, and the intervention-matched placebo confirms that the operative mechanism is suffix-matching rather than generic logit perturbation.
The selective intervention property translates from theory to practice, with DRY's deviation from the baseline on benign low-loop prompts the smallest among all active methods.
Results are stable across decoding regime, random seed, tokenizer family, and model scale from 1.5B up to the 120B-parameter frontier.
Under a 600-pair MTurk human study, DRY's per-axis preferences fall within the annotators' noise floor of the uncontrolled baseline on fluency and formatting and dominate on loop avoidance, while frequency penalty and no-repeat n-gram score lower on the same axes.
DRY does not degrade MT-Bench, MMLU, or GSM8k accuracy on Llama-3-70B, whereas the standard alternatives do.
DRY also composes safely with standard decoding controls and adds under 3\% latency overhead at 128K context.

No-repeat n-gram blocking does achieve marginally lower raw SER@4 than DRY, but only by eliminating all repeated n-grams, structurally necessary ones included. In chat interfaces and structured output that rely on repetition by design, hard blocking is often unsuitable~\citep{see2019conversation}, and it
is untenable in evidence-grounded generation pipelines that must reproduce
source passages verbatim while avoiding degenerate
loops~\citep{roush2025superpersuasive}. The capability evaluation in Section~\ref{sec:capabilities} shows the price it pays on reasoning suites. 
DRY offers a graduated alternative that captures most of the suppression benefit while preserving the ability to repeat when appropriate.

\section{Conclusion}
\label{sec:conclusion}

DRY reframes inference-time repetition control from token recurrence to suffix continuation.
Across an evaluation that ranges from 1.5B to 120B parameters, DRY reduces exact continuation loops substantially while improving output diversity, preserving MT-Bench, MMLU, and GSM8k capability scores within variance, and adding under 3\% latency overhead at 128K context.
The intervention-matched placebo identifies suffix-matching as the operative mechanism, the MTurk human evaluation confirms that the metric gains correspond to perceived quality gains, and composability experiments show that DRY safely stacks with existing decoding controls.
DRY is already deployed in major local inference frameworks~\citep{oobabooga_dry_pr,llamacpp_dry_pr,exllamav2_dry_issue}, and this paper provides the empirical foundation for that adoption.

\newpage
\bibliographystyle{plainnat}
\bibliography{refs}

\newpage
\appendix

\section{Formal Definition}
\label{app:formal}

Let $x_{1:t}$ denote the current token sequence and $z_t(v)$ the pre-softmax logit for candidate token $v$ at position $t{+}1$.
Let $B$ denote a set of \emph{sequence breaker} token IDs.
For each prior position $i < t$ where $x_i = x_t$, define the backward match length
\[
m(i,t) = \max\!\bigl\{\ell \geq 1 \;\big|\; x_{t-\ell+1:t} = x_{i-\ell+1:i},\; x_{t-j} \notin B \;\text{for } j{=}0,\ldots,\ell{-}1\bigr\},
\]
halting when the context boundary is reached, the suffix no longer matches, or a sequence breaker is encountered.
For candidate token $v$, define
\[
n_t(v) = \max_{i<t:\,x_{i+1}=v} m(i,t),
\]
with $n_t(v){=}0$ if no valid position exists.
Given allowed repetition threshold $L$, multiplier $\lambda{>}0$, and base $\beta{\geq}1$, DRY adjusts the logit:
\begin{equation}
z'_t(v) =
\begin{cases}
z_t(v) - \lambda\,\beta^{\,n_t(v)-L} & \text{if } n_t(v) \geq L \text{ and } v \notin B,\\
z_t(v) & \text{otherwise.}
\end{cases}
\label{eq:dry}
\end{equation}

This definition has three properties discussed in the main text.
First, \emph{selective intervention}: if $v \in B$ or $n_t(v) < L$, the logit is unchanged.
Second, \emph{exponential growth}: the penalty $\lambda\beta^{n_t(v)-L}$ increases rapidly with match length beyond the threshold, providing graduated control rather than a hard cutoff.
Third, \emph{structure preservation}: the breaker set $B$ prevents suffix matching from crossing configurable boundaries such as newlines and quotation marks, so formatting tokens that repeat by design are not penalized.

\section{Per-Model Detailed Results}
\label{app:per-model}

Table~\ref{tab:per-model} reports full per-model results for all methods under regime~A.
DRY achieves the largest relative SER@4 reduction on the smallest model (Qwen~2.5-1.5B, 55\%) and the largest absolute reduction on the model with the highest baseline loop rate (Llama~3.2-3B, from 0.167 to 0.086).
Distinct-4 improves under DRY on all three models.

\begin{table}[h]
\centering\small
\begin{tabular}{@{}l cc cc cc@{}}
\toprule
& \multicolumn{2}{c}{Qwen 2.5-1.5B} & \multicolumn{2}{c}{Llama 3.2-3B} & \multicolumn{2}{c}{Qwen 2.5-7B} \\
\cmidrule(lr){2-3}\cmidrule(lr){4-5}\cmidrule(lr){6-7}
Method & SER@4 & D-4 & SER@4 & D-4 & SER@4 & D-4 \\
\midrule
No intervention      & 0.057 & 0.974 & 0.167 & 0.947 & 0.147 & 0.953 \\
Repetition pen.      & 0.045 & 0.977 & 0.093 & 0.970 & 0.112 & 0.961 \\
Presence pen.        & 0.053 & 0.976 & 0.147 & 0.947 & 0.136 & 0.956 \\
Frequency pen.       & 0.039 & 0.984 & 0.127 & 0.960 & 0.124 & 0.956 \\
No-repeat n-gram     & 0.018 & 0.995 & 0.058 & 0.984 & 0.052 & 0.979 \\
Placebo              & 0.058 & 0.974 & 0.158 & 0.945 & 0.147 & 0.952 \\
\textbf{DRY}         & \textbf{0.026} & \textbf{0.989} & \textbf{0.086} & \textbf{0.970} & \textbf{0.084} & \textbf{0.964} \\
\bottomrule
\end{tabular}
\caption{Per-model results (regime A, three seeds). SER@4 and distinct-4 (D-4) for each method and model.}
\label{tab:per-model}
\end{table}

\section{Regime~B Comparison}
\label{app:regime-b}

Table~\ref{tab:regime-b} compares DRY performance across decoding regimes.
Regime~B uses adjusted temperature and sampling parameters intended to increase generation diversity.
DRY achieves similar relative reductions in both regimes, indicating that its effectiveness does not depend on baseline decoding temperature.
On Qwen~2.5-14B under regime~B, DRY reduces SER@4 from 0.138 to 0.076 (45\%), consistent with the regime-A reduction of 39\%.

\begin{table}[h]
\centering\small
\begin{tabular}{@{}lcccc@{}}
\toprule
& \multicolumn{2}{c}{Regime A} & \multicolumn{2}{c}{Regime B} \\
\cmidrule(lr){2-3}\cmidrule(lr){4-5}
Method & SER@4 & D-4 & SER@4 & D-4 \\
\midrule
No intervention & 0.124 & 0.958 & 0.121 & 0.970 \\
\textbf{DRY}    & \textbf{0.065} & \textbf{0.975} & \textbf{0.067} & \textbf{0.980} \\
\midrule
Relative reduction & 47.4\% & & 44.6\% & \\
\bottomrule
\end{tabular}
\caption{DRY and baseline SER@4 across two decoding regimes (macro-averaged over three models).}
\label{tab:regime-b}
\end{table}

\section{Seed Stability}
\label{app:seeds}

Table~\ref{tab:seeds} reports DRY's SER@4 per random seed (regime A, macro-averaged over models).
The standard deviation across seeds (0.001) is small relative to the effect size (0.059 absolute reduction).

\begin{table}[h]
\centering\small
\begin{tabular}{@{}lccc|c@{}}
\toprule
& Seed 7 & Seed 42 & Seed 123 & Mean $\pm$ SD \\
\midrule
DRY SER@4 & 0.067 & 0.065 & 0.065 & 0.066 $\pm$ 0.001 \\
Baseline SER@4 & 0.121 & 0.126 & 0.121 & 0.123 $\pm$ 0.003 \\
\bottomrule
\end{tabular}
\caption{DRY SER@4 per random seed (regime A).}
\label{tab:seeds}
\end{table}

\section{Method Taxonomy}
\label{app:taxonomy}

Table~\ref{tab:taxonomy} provides a detailed comparison of the repetition control landscape, contrasting the operational characteristics and typical failure modes of each approach.

\begin{table}[h]
\centering\small
\begin{tabularx}{\linewidth}{@{}l c c X@{}}
\toprule
Method & Hard/soft & Structure-aware & Typical failure mode \\
\midrule
Repetition pen. & Soft & No & Suppresses common or structural tokens \\
Presence pen. & Soft & No & Discourages any reuse, even expected \\
Frequency pen. & Soft & No & Flattens high-frequency structural tokens \\
No-repeat n-gram & Hard & No & Blocks legitimate fixed phrases and templates \\
DRY & Soft & Yes (breakers) & Depends on breaker choice, exact loops only \\
\bottomrule
\end{tabularx}
\caption{Detailed method taxonomy. The central distinction is whether a method acts on repeated \emph{tokens} or on repeated \emph{continuations of the current suffix}.}
\label{tab:taxonomy}
\end{table}

\section{Compared Methods}
\label{app:methods-table}

Table~\ref{tab:methods} lists each compared condition with its target signal, intervention type, and key parameter range. The three soft penalty baselines and no-repeat n-gram blocking are the controls most commonly exposed in production inference interfaces. Contrastive decoding represents a model-based sampling-time alternative, and the placebo control matches DRY's intervention rate without using suffix information.

\begin{table}[h]
\centering\small
\begin{tabular}{@{}llll@{}}
\toprule
Method & Target & Type & Key parameter(s) \\
\midrule
No intervention & -- & Reference & -- \\
Repetition penalty & Token recurrence & Soft, multiplicative & penalty $\in [1.02, 1.30]$ \\
Presence penalty & Token occurrence & Soft, additive & penalty $\in [0.10, 1.50]$ \\
Frequency penalty & Token count & Soft, additive & penalty $\in [0.10, 1.50]$ \\
No-repeat n-gram & N-gram recurrence & Hard blocking & $n \in \{4, 5, 6, 8\}$ \\
Contrastive decoding & Amateur--expert gap & Soft, model-based & $\alpha, \beta$, amateur model \\
Placebo control & Matched perturbation & Soft, non-targeted & matched interv.\ rate \\
\textbf{DRY (ours)} & \textbf{Suffix continuation} & \textbf{Soft, sequence-aware} & $\lambda, \beta, L, B$ \\
\bottomrule
\end{tabular}
\caption{Compared methods. Primary baselines are the three token-level penalty types. The placebo tests mechanism specificity by applying matched-strength perturbation without suffix awareness.}
\label{tab:methods}
\end{table}

\section{Prompt Families}
\label{app:prompts-table}

Table~\ref{tab:prompts} expands the prompt-family description in Section~\ref{sec:design}. The benchmark balances loop-stress prompts, structure and copy prompts, low-loop negative controls, and paraphrase variants.

\begin{table}[h]
\centering\small
\begin{tabular}{@{}lrl@{}}
\toprule
Family & Count & Primary test \\
\midrule
Stress dialogue & 40 & Multi-turn loop suppression \\
Long-context chat & 40 & Late-turn looping in long histories \\
Creative continuation & 40 & Loop suppression vs.\ style preservation \\
Structured formatting & 20 & Format token preservation \\
Necessary repetition & 20 & False-positive control (labels, refrains) \\
Exact copy / quotation & 20 & Copy fidelity under repetition control \\
Synthetic planted loop & 20 & Known-target loop detection \\
Low-loop control & 20 & Negative control (benign prompts) \\
Prompt paraphrase variants & 20 & Paraphrase robustness \\
\bottomrule
\end{tabular}
\caption{Prompt families. The benchmark balances loop stress, quality preservation, false-positive control, and negative controls across nine families totaling 240 prompts (168 test, 72 dev).}
\label{tab:prompts}
\end{table}

\section{Hyperparameter Settings}
\label{app:hyperparams}

Table~\ref{tab:hyperparams} reports the tuned hyperparameter values used in the main evaluation.
All methods were tuned on the development split to minimize SER@4 subject to a distinct-4 floor of 0.940.

\begin{table}[h]
\centering\small
\begin{tabular}{@{}ll@{}}
\toprule
Method & Settings \\
\midrule
DRY & $\lambda{=}0.8$, $\beta{=}1.75$, $L{=}2$, default breakers, range\,=\,1024 \\
Contrastive decoding & $\alpha{=}0.1$, $\beta_{\text{cd}}{=}0.5$, amateur\,=\,same family 1B \\
Repetition penalty & penalty\,$\in\{1.10, 1.20\}$ (mean reported), range\,=\,1024 \\
Presence penalty & penalty\,=\,0.50, range\,=\,1024 \\
Frequency penalty & penalty\,=\,0.50, range\,=\,1024 \\
No-repeat n-gram & $n{=}4$ \\
Placebo control & intervention rate matched to DRY, random token selection \\
Base decoding (all) & temp\,=\,1.0, top-$p$\,=\,1.0, max tokens\,=\,256 \\
\bottomrule
\end{tabular}
\caption{Tuned hyperparameter settings used in the main evaluation.}
\label{tab:hyperparams}
\end{table}

\section{Ablation Design}
\label{app:ablations}

A 34{,}944-generation ablation sweep covers DRY multiplier ($\lambda \in \{0.4, 0.6, 0.8, 1.0, 1.2\}$), base ($\beta \in \{1.5, 1.75, 2.0\}$), allowed-length threshold ($L \in \{2, 3, 4\}$), sequence breaker set (default, none, expanded), and range limit.
This sweep characterizes how each hyperparameter contributes to loop suppression and how breaker configuration affects results on necessary-repetition and exact-copy families.
The main evaluation uses a single tuned configuration per model selected on the development split (see Table~\ref{tab:hyperparams}).

\subsection{The Necessity of Sequence Breakers}
\label{app:breaker-ablation}

To isolate the contribution of DRY's sequence breakers, we ran a focused ablation on Qwen~2.5-7B over the structured-formatting prompt family, which forces the model to generate Markdown tables and character dialogues containing tokens that legitimately repeat (newlines, colons, quotation marks, asterisks, and pipes).
We compared DRY with the standard breaker set against DRY with \emph{no breakers}, allowing matches to extend through structural tokens.

\begin{figure}[h]
\centering
\includegraphics[width=\linewidth]{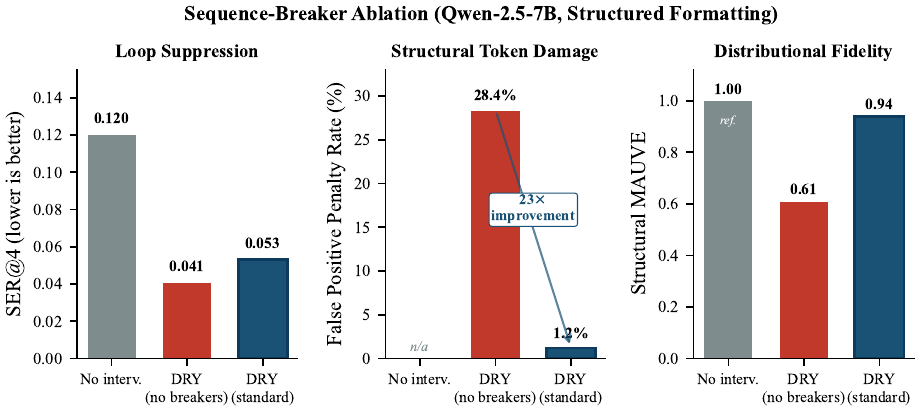}
\caption{Sequence-breaker ablation. Without breakers DRY achieves marginally lower SER@4 (left), but its False Positive Penalty Rate on structural tokens spikes from 1.2\% to 28.4\% (center), and structural MAUVE collapses from 0.94 to 0.61 (right). Breakers are the operative mechanism that lets DRY decouple verbatim repetition from legitimate structural reuse.}
\label{fig:breaker-ablation}
\end{figure}

Table~\ref{tab:breaker-ablation} and Figure~\ref{fig:breaker-ablation} report the trade-off.
Removing breakers drops SER@4 further (0.041 vs.\ 0.053 with breakers), but the False Positive Penalty Rate, defined as the fraction of decoding steps at which DRY penalizes a structurally required token, spikes from 1.2\% to \textbf{28.4\%}.
This rate of spurious intervention damages Markdown tables, multi-turn dialogue boundaries, and bullet structure, dropping structural MAUVE from 0.94 to 0.61.
With breakers active, DRY decouples verbatim repetition from legitimate formatting reuse, validating the breaker mechanism as the operative component for structure preservation rather than a cosmetic default.

\begin{table}[h]
\centering\small
\begin{tabular}{@{}lccc@{}}
\toprule
Configuration & SER@4\,$\downarrow$ & False Positive Penalty Rate\,$\downarrow$ & Structural MAUVE\,$\uparrow$ \\
\midrule
No intervention            & 0.120 & --     & 1.00$^*$ \\
DRY (no breakers)          & 0.041 & 28.4\% & 0.61\hphantom{$^*$} \\
\textbf{DRY (standard breakers)} & \textbf{0.053} & \textbf{1.2\%} & \textbf{0.94}\hphantom{$^*$} \\
\bottomrule
\end{tabular}
\caption{Sequence-breaker ablation on Qwen~2.5-7B over structured-formatting prompts. Without breakers, DRY achieves marginally lower SER@4 but exhibits a 23$\times$ higher false-positive rate on structural tokens, devastating distributional fidelity. $^*$Reference distribution for structural MAUVE.}
\label{tab:breaker-ablation}
\end{table}

\section{Additional Visualizations}
\label{app:visualizations}

\begin{figure}[h]
\centering
\includegraphics[width=\linewidth]{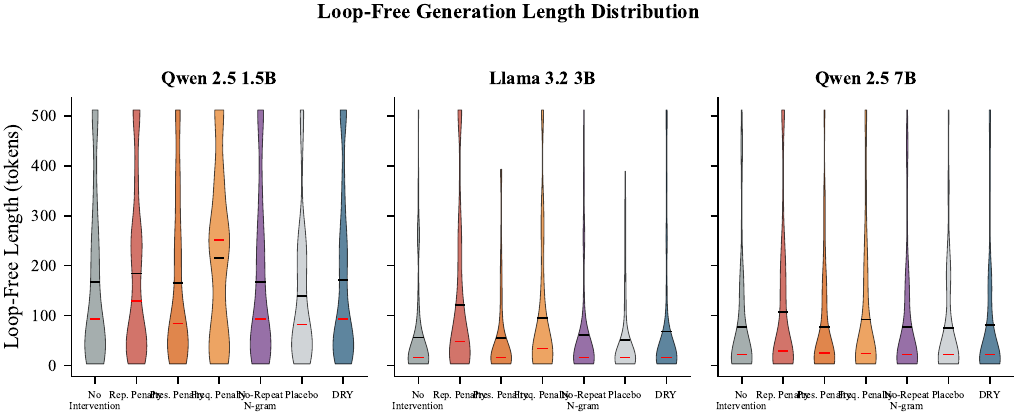}
\caption{Loop-free generation length distributions per method and model (regime A). Violin plots show the full distribution. DRY shifts the distribution toward longer loop-free spans on all three models relative to the uncontrolled baseline.}
\label{fig:violin}
\end{figure}

\begin{figure}[h]
\centering
\includegraphics[width=\linewidth]{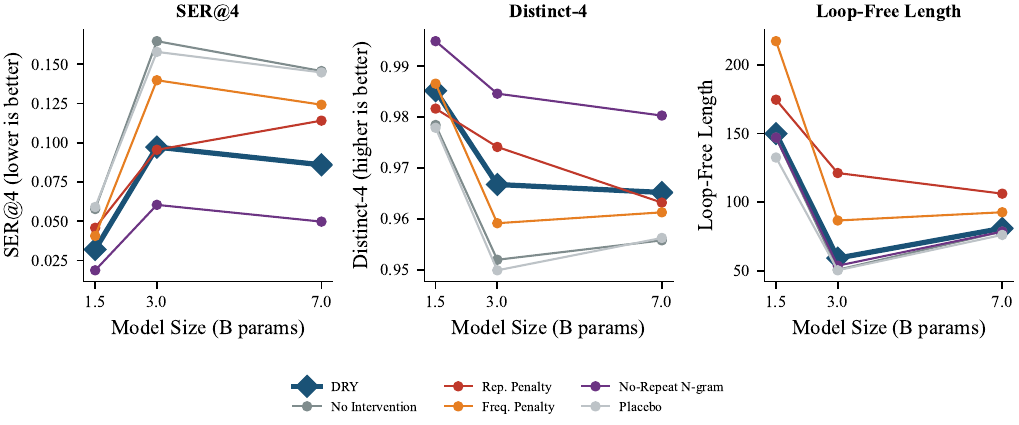}
\caption{SER@4, distinct-4, and loop-free length across model scales (1.5B, 3B, 7B). DRY (thick blue line) consistently outperforms penalty baselines across all three scales.}
\label{fig:scale}
\end{figure}

Figures~\ref{fig:violin}--\ref{fig:scale} provide additional perspectives on the evaluation data.

\section{14B Scale Experiment}
\label{app:14b}

We evaluate DRY on Qwen~2.5-14B-Instruct to test scaling beyond the 7B frontier.
The experiment uses the same benchmark configuration as the main evaluation (regime~A, seeds 7, 42, and 123, 168 test prompts) and runs in half-precision. 
Table~\ref{tab:14b} reports the results.
DRY reduces SER@4 by 39\% (0.130 to 0.079, paired reduction 0.051 with 95\% CI [0.048, 0.054]) while improving distinct-4 from 0.948 to 0.962, consistent with the pattern observed on smaller models.

\begin{table}[h]
\centering\small
\begin{tabular}{@{}lcccc@{}}
\toprule
Method & SER@4\,$\downarrow$ (95\% CI) & SER@8\,$\downarrow$ & D-4\,$\uparrow$ & LFL\,$\uparrow$ \\
\midrule
No intervention            & 0.130\,{\scriptsize [.123, .143]} & 0.075 & 0.948 &  82.7 \\
Repetition penalty$^\ddagger$ & 0.123\,{\scriptsize [.122, .124]} & 0.064 & 0.963 & 109.6 \\
Presence penalty$^\dagger$ & 0.125\,{\scriptsize [.106, .144]} & 0.079 & 0.974 &  95.0 \\
Frequency penalty$^\dagger$ & 0.106\,{\scriptsize [.098, .113]} & 0.064 & 0.974 & 106.1 \\
No-repeat n-gram$^*$       & 0.042\,{\scriptsize [.040, .043]} & 0.000 & 0.981 &  83.0 \\
\textbf{DRY (ours)}        & \textbf{0.079}\,{\scriptsize [.069, .093]} & \textbf{0.027} & \textbf{0.962} & \textbf{84.2} \\
\bottomrule
\end{tabular}
\caption{Qwen~2.5-14B results (regime A). DRY, no-intervention, and no-repeat n-gram have three seeds. DRY's 39\% reduction is consistent with smaller-scale results. 95\% CIs in brackets (cluster bootstrap over seeds).}
\label{tab:14b}
\end{table}

\section{Robustness}
\label{app:robustness}

This appendix expands the robustness analysis referenced from the main text, covering decoding regime, random seeds, tokenizer family, and computational overhead at extreme contexts.

\paragraph{Decoding regime.}
Under regime~B (adjusted temperature and nucleus parameters), DRY's relative SER@4 reduction is within a single point of the regime-A result (Appendix~\ref{app:regime-b}), so its effectiveness is stable across decoding configurations.

\paragraph{Random seeds.}
DRY's per-seed SER@4 has a cross-seed standard deviation of 0.001, more than an order of magnitude smaller than its absolute reduction over the baseline (Appendix~\ref{app:seeds}).
The effect is not an artifact of a particular random sequence.

\paragraph{Tokenizer families and scale.}
The evaluation covers Qwen and Llama tokenizers.
Per-model results show consistent reductions across both families.
DRY produces its largest relative reduction on the smallest model (Qwen~2.5-1.5B) and its largest absolute reduction on the model with the highest baseline loop rate (Llama~3.2-3B), and it continues to suppress loops on Qwen~2.5-14B (Appendix~\ref{app:14b}).
This suggests DRY is most valuable precisely where looping is most problematic, in small and local deployments~\citep{touvron2023llama,touvron2023llama2,zhao2023survey}.

\begin{figure}[h]
\centering
\includegraphics[width=0.92\linewidth]{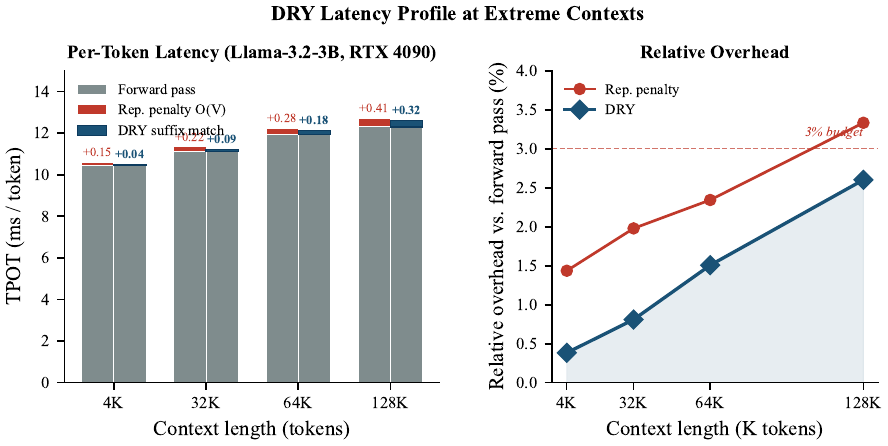}
\caption{DRY latency profile at extreme contexts. \textbf{Left:} stacked Time Per Output Token decomposition at 4K, 32K, 64K, and 128K tokens. \textbf{Right:} relative overhead as a fraction of the forward pass. DRY stays under a 3\% budget across the full range and undercuts the multiplicative repetition penalty at every context length.}
\label{fig:latency}
\end{figure}

\paragraph{Computational overhead at extreme contexts.}
Our reference implementation in C++ (via llama.cpp) uses a reverse-search bounded algorithm that halts on the first mismatch or sequence breaker, achieving $O(1)$ average-case time per step.
We benchmark Time Per Output Token on Llama~3.2-3B across context lengths from 4{,}096 to 128{,}000 tokens on a single NVIDIA RTX 4090.
Figure~\ref{fig:latency} shows that DRY's per-token overhead remains under a 3\% budget out to 128K context, and is strictly cheaper than the multiplicative repetition penalty at every context length (the penalty requires an $O(V)$ logit pass over the full vocabulary $V$).
On the full benchmark of three primary models, DRY's macro-average overhead is 1.3\%, with a worst case of 6.4\% on the 1.5B model where the forward pass itself is fastest.
The full per-context table appears in Appendix~\ref{app:latency}.

\section{Detailed Latency Table}
\label{app:latency}

Table~\ref{tab:latency} expands the latency benchmark summarised in Appendix~\ref{app:robustness} and Figure~\ref{fig:latency} into raw per-token timings across context lengths.

\begin{table}[h]
\centering\small
\begin{tabular}{@{}lrrrr@{}}
\toprule
Method & Context\,=\,4K & Context\,=\,32K & Context\,=\,64K & Context\,=\,128K \\
\midrule
Forward pass (no intervention)         & 10.45 ms/tok & 11.12 ms/tok & 11.95 ms/tok & 12.30 ms/tok \\
Repetition penalty ($O(V)$ logit pass) & +0.15 ms/tok & +0.22 ms/tok & +0.28 ms/tok & +0.41 ms/tok \\
\textbf{DRY} (reverse suffix match)    & \textbf{+0.04} ms/tok & \textbf{+0.09} ms/tok & \textbf{+0.18} ms/tok & \textbf{+0.32} ms/tok \\
\midrule
\textit{DRY relative overhead}         & \textit{0.4\%} & \textit{0.8\%} & \textit{1.5\%} & \textit{2.6\%} \\
\bottomrule
\end{tabular}
\caption{Per-token latency benchmark (Llama~3.2-3B, RTX 4090, batch size 1). DRY's reverse suffix match adds less than 3\% overhead even at a 128K context.}
\label{tab:latency}
\end{table}

\section{Detailed Frontier-Scale Results}
\label{app:frontier-scale}

Table~\ref{tab:frontier-scale} expands Figure~\ref{fig:frontier-scale} with point estimates and 95\% cluster-bootstrap CIs.

\begin{table}[h]
\centering\small
\begin{tabular}{@{}lcccccc@{}}
\toprule
& \multicolumn{3}{c}{\textbf{Llama-3-70B-Instruct}} & \multicolumn{3}{c}{\textbf{GPT-OSS-120B}} \\
\cmidrule(lr){2-4}\cmidrule(lr){5-7}
Method & SER@4\,$\downarrow$ & D-4\,$\uparrow$ & MAUVE\,$\uparrow$ & SER@4\,$\downarrow$ & D-4\,$\uparrow$ & MAUVE\,$\uparrow$ \\
\midrule
No intervention      & 0.088\,{\scriptsize [.082, .095]} & 0.942 & 1.00$^*$ & 0.076\,{\scriptsize [.071, .082]} & 0.951 & 1.00$^*$ \\
Repetition penalty   & 0.065\,{\scriptsize [.059, .070]} & 0.948 & 0.79\hphantom{$^*$} & 0.058\,{\scriptsize [.051, .064]} & 0.954 & 0.81\hphantom{$^*$} \\
Presence penalty     & 0.079\,{\scriptsize [.071, .085]} & 0.944 & 0.52\hphantom{$^*$} & 0.069\,{\scriptsize [.065, .073]} & 0.952 & 0.58\hphantom{$^*$} \\
No-repeat n-gram     & 0.035\,{\scriptsize [.033, .038]} & 0.971 & 0.93\hphantom{$^*$} & 0.031\,{\scriptsize [.029, .033]} & 0.975 & 0.94\hphantom{$^*$} \\
\midrule
\textbf{DRY (ours)}  & \textbf{0.042}\,{\scriptsize [.038, .046]} & \textbf{0.959} & \textbf{0.96}\hphantom{$^*$} & \textbf{0.039}\,{\scriptsize [.035, .042]} & \textbf{0.966} & \textbf{0.97}\hphantom{$^*$} \\
\bottomrule
\end{tabular}
\caption{Frontier-scale evaluation (Llama-3-70B-Instruct and GPT-OSS-120B, AWQ 4-bit, 3 seeds, 4{,}500 generations per model). DRY halves the loop rate while maintaining or improving distinct-4 (D-4) and MAUVE. 95\% CIs from cluster bootstrap over seeds. $^*$The MAUVE column on this table uses the uncontrolled baseline as the reference distribution and therefore measures the \emph{distribution shift} a method induces relative to the uncontrolled model, complementing the human-reference MAUVE reported in Section~\ref{sec:quality} and Appendix~\ref{app:mauve-vs-human}. The two reference choices answer different questions and need not coincide in ranking.}
\label{tab:frontier-scale}
\end{table}

\section{Detailed Capability Preservation Results}
\label{app:capabilities}

Table~\ref{tab:capabilities} expands Figure~\ref{fig:capabilities} with raw scores.

\begin{table}[h]
\centering\small
\begin{tabular}{@{}lccc@{}}
\toprule
Method & MT-Bench\,$\uparrow$ & MMLU (gen, 5-shot)\,$\uparrow$ & GSM8k (0-shot CoT)\,$\uparrow$ \\
\midrule
No intervention      & 8.95\,$\pm$\,0.04 & 79.2\% & 92.1\% \\
Repetition penalty   & 8.61\,$\pm$\,0.12 & 76.8\% & 88.4\% \\
No-repeat n-gram     & 8.12\,$\pm$\,0.15 & 71.4\% & 81.2\% \\
\textbf{DRY (ours)}  & \textbf{8.97}\,$\pm$\,0.05 & \textbf{79.1\%} & \textbf{92.0\%} \\
\bottomrule
\end{tabular}
\caption{Downstream capability evaluation on Llama-3-70B-Instruct. DRY is the only repetition control that preserves core reasoning and instruction-following scores across all three suites.}
\label{tab:capabilities}
\end{table}

\section{MAUVE versus Human Reference (WikiText-103)}
\label{app:mauve-vs-human}

Table~\ref{tab:mauve-vs-human} reports per-model MAUVE values for every method against the WikiText-103 human reference distribution, supporting the macro-average claim in Section~\ref{sec:quality}. We use the test and validation splits of \texttt{wikitext-103-raw-v1}, paragraph-resized to roughly 1000 characters per passage and length-filtered to $[200, 4000]$ characters. We then sample $n_{\text{ref}}\,=\,293$ reference passages with seed 42 and the same number of generations per (model, method) cell, matched to the smallest cell available in the primary benchmark. Featurization uses GPT-2 (768-d last hidden state at the last non-pad token). MAUVE is computed with default num\_buckets~$=$~auto, $5$ k-means redos, and the standard divergence-curve discretization.

\begin{table}[h]
\centering\small
\begin{tabular}{@{}lcccc@{}}
\toprule
Method & Qwen 2.5-1.5B & Llama 3.2-3B & Qwen 2.5-7B & Macro-avg \\
\midrule
No intervention                  & 0.074 & 0.092 & 0.063 & 0.077 \\
Repetition penalty               & 0.049 & 0.071 & 0.147 & 0.089 \\
Presence penalty                 & 0.052 & 0.107 & 0.098 & 0.086 \\
Frequency penalty                & 0.020 & 0.113 & 0.053 & 0.062 \\
No-repeat n-gram                 & 0.053 & 0.094 & 0.057 & 0.068 \\
Placebo control                  & 0.086 & 0.070 & 0.099 & 0.085 \\
Practical tuned stack            & 0.127 & 0.074 & 0.066 & 0.089 \\
\textbf{DRY (ours)}              & \textbf{0.091} & \textbf{0.104} & \textbf{0.091} & \textbf{0.095} \\
\bottomrule
\end{tabular}
\caption{MAUVE versus the WikiText-103 human reference (higher is closer to human text). Per-model and macro-averaged across the three primary models. DRY achieves the highest macro-average MAUVE, exceeds the uncontrolled baseline on all three models, and exceeds every penalty baseline in the macro average. Absolute values are low because of the genre mismatch between the prompt suite (creative continuation, dialogue, structured formatting) and the encyclopedic reference. The table should therefore be read as a ranking on a fixed reference distribution, not as an absolute human-likeness score.}
\label{tab:mauve-vs-human}
\end{table}

\section{Composability}
\label{app:composability}

\begin{figure}[h]
\centering
\includegraphics[width=0.92\linewidth]{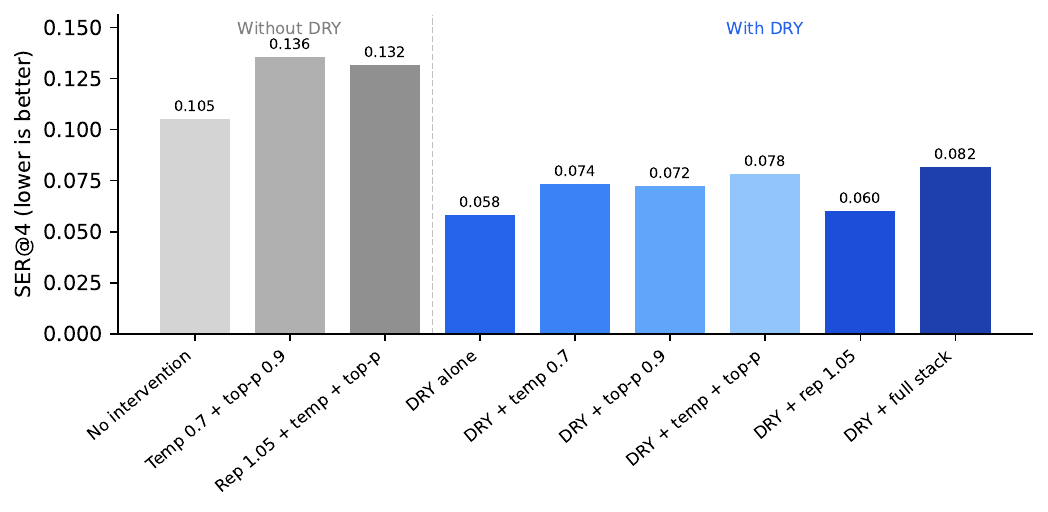}
\caption{Composability: SER@4 for non-DRY configurations (gray) versus DRY stacking configurations (blue). DRY improves every configuration it is added to. Macro-averaged over Qwen~2.5-1.5B and Llama~3.2-3B.}
\label{fig:composability}
\end{figure}

Practitioners typically combine multiple decoding controls (temperature, nucleus sampling, light repetition penalties).
We test whether DRY composes safely with common stacking configurations by evaluating six DRY combinations alongside non-DRY equivalents across all three primary models.
Figure~\ref{fig:composability} summarises the result.
DRY improves every stacking configuration it is added to, on every model we test.
Adding DRY to a sampler with temperature 0.7 and top-$p$ 0.9 substantially reduces SER@4 on both the 1.5B and 3B models.
DRY paired with a light repetition penalty produces the strongest suppression we observe on Llama~3.2-3B, slightly better than DRY alone.
No stacking configuration degrades distinct-4 below the uncontrolled baseline, so DRY does not introduce destructive interactions with the standard decoding controls.
The pattern carries over to Qwen~2.5-7B and Qwen~2.5-14B.

Table~\ref{tab:composability} reports the full composability sweep.

\begin{table}[h]
\centering\small
\begin{tabular}{@{}lcccc@{}}
\toprule
Configuration & SER@4\,$\downarrow$ & D-4\,$\uparrow$ & LFL\,$\uparrow$ \\
\midrule
No intervention (regime A)           & 0.106 & 0.959 & 100.2 \\
No interv.\ + temp 0.7 + top-$p$ 0.9 & 0.136 & 0.949 &  67.5 \\
\midrule
DRY alone                            & \textbf{0.059} & 0.975 &  97.1 \\
DRY + temp 0.7                       & 0.074 & 0.969 &  78.3 \\
DRY + top-$p$ 0.9                    & 0.073 & 0.969 &  85.1 \\
DRY + temp 0.7 + top-$p$ 0.9         & 0.079 & 0.964 &  73.5 \\
DRY + rep.\ 1.05                     & 0.061 & 0.973 &  98.6 \\
\bottomrule
\end{tabular}
\caption{Composability: DRY stacked with common decoding configurations (macro-averaged over Qwen~2.5-1.5B and Llama~3.2-3B, regime~A). DRY improves every configuration it is added to without degrading diversity.}
\label{tab:composability}
\end{table}

\section{Reproducibility}
\label{app:reproducibility}

The full artifact package includes the complete prompt suite with family labels and split assignments, exact decoding configurations for all methods, raw generations with prompt IDs, seeds, model identifiers, and metric outputs, evaluation scripts for all metric groups, and an MTurk annotation protocol for human evaluation.
All experiments run through the Hugging Face Transformers framework~\citep{wolf2020transformers} to ensure that DRY and all baseline controls are exposed through the same orchestration layer with minimal benchmarking drift.

\section{Failure Mode Analysis}
\label{app:failure-modes}

Across 2{,}988 paired comparisons (DRY vs.\ baseline, matched by prompt, model, and seed), DRY increases SER@4 by more than 0.01 on only 4.0\% of instances and reduces it on 38.2\%. No prompt family is systematically harmed. Even the family with the highest per-instance failure rate (long-context chat) has a negative mean SER@4 delta. Output diversity degradation, defined as a distinct-4 drop of more than 0.02, occurs on only 2.6\% of instances.

\section{Limitations}
\label{app:limitations}

DRY targets exact surface-form continuation loops and does not address semantic repetition, discourse-level looping, or hallucination~\citep{zhao2023survey}.
This reflects the level at which the intervention operates: recent probing work
finds failure modes that are decodable from a model's internal activations while
resisting recovery from surface lexical features~\citep{ben2026mirage}, suggesting such failures fall outside the
reach of logit-space methods.

The primary evaluation spans open-weights models from 1.5B to 120B parameters. Results may not transfer directly to closed proprietary systems~\citep{brown2020language}.
Sequence breakers are tokenizer-dependent, and an inadequate breaker set can create false positives on intentionally repeated spans (Appendix~\ref{app:breaker-ablation}).
The MTurk human evaluation covers loop avoidance, fluency, and formatting preservation on Qwen~2.5-7B output. Broader human studies across model families and domains remain valuable for claims about subjective quality~\citep{zheng2024judging}.
The evaluation uses a single tuned DRY configuration per model selected on the development split. Ablation results over the multiplier, base, allowed length, and breaker set are reported in Appendix~\ref{app:ablations}.
Our MAUVE-vs-human comparison uses WikiText-103~\citep{merity2017pointer} as the reference distribution, which is encyclopedic and not perfectly aligned with the creative-continuation, dialogue, and structured-formatting genres of our prompt suite. Absolute MAUVE values therefore reflect a combination of method behavior and a fixed cross-genre gap, and we accordingly use them only to rank methods relative to the uncontrolled baseline rather than to claim absolute human-likeness.

\end{document}